\pdfoutput=1

\documentclass[dvipsnames]{article}
\usepackage{arxiv}

\usepackage[boxed,ruled,vlined,linesnumbered]{algorithm2e}
\DontPrintSemicolon
\SetKwProg{Fn}{function}{}{}
\SetKwFunction{FnGetAction}{GetAction}
\SetKwFunction{FnDelegatePolicy}{DelegatePolicy}
\SetKwFunction{FnRestartArm}{RestartArm}
\SetKwFunction{FnStep}{Step}
\SetKwComment{Comment}{$\triangleright$\ }{}
\SetKwInput{KwInit}{Initialise}

\usepackage{etoolbox}
\makeatletter
\patchcmd{\@algocf@start}%
{-1.5em}%
{0pt}%
{}{}%
\makeatother

\usepackage{wrapfig}
\usepackage{minibox}

\let\labelindent\relax
\usepackage[inline]{enumitem}

\usepackage{subcaption}

\usepackage[activate={true,nocompatibility},final,tracking=true,kerning=true,spacing=true,factor=1100,stretch=10,shrink=10]{microtype}

\microtypecontext{spacing=nonfrench}

\usepackage{booktabs}
\usepackage{multirow}

\usepackage{amssymb}                                        %
\usepackage{mathtools}
\usepackage{braket}                                         %
\DeclareMathOperator*{\argmax}{arg\,max}

\newtheorem{definition}{Definition}

\usepackage{graphicx}                                       %
\usepackage{tikz}

\usepackage{bm}
\usepackage{dsfont}
\usepackage{xifthen}
\usepackage{stackengine} %
\usepackage{url}

\usepackage{cleveref}                                        %
\crefname{assumption}{assumption}{assumptions}
\crefname{problem}{problem}{problems}
\crefname{algorithm}{Alg.}{Algs.}
\Crefname{algorithm}{Algorithm}{Algorithms}
\crefname{figure}{Figure}{Figs.} %
\crefformat{equation}{(#2#1#3)}
\crefrangeformat{equation}{(#3#1#4) to~(#5#2#6)}
\crefmultiformat{equation}{(#2#1#3)}%
{ and~(#2#1#3)}{, (#2#1#3)}{ and~(#2#1#3)}

\crefformat{footnote}{#2\footnotemark[#1]#3}

\usepackage[boxed,ruled,vlined,linesnumbered]{algorithm2e}
\DontPrintSemicolon
\SetKwProg{Fn}{function}{}{}
\SetKwFunction{FnSampleFree}{SampleFree}
\SetKwFunction{FnRestartArm}{RestartArm}
\SetKwFunction{FnPickArm}{PickArm}
\SetKwFunction{FnRewire}{Rewire}
\SetKwFunction{FnNearest}{Nearest}
\SetKwFunction{FnRestartArm}{RestartArm}
\SetKwComment{Comment}{$\triangleright$\ }{}
\SetKwInput{KwInit}{Initialise}

\SetCommentSty{commentfont}

\definecolor{amethyst}{rgb}{0.6, 0.4, 0.8}
\definecolor{alizarin}{rgb}{0.82, 0.1, 0.26}
\definecolor{ashgrey}{rgb}{0.43, 0.5, 0.5}
\definecolor{yellow}{rgb}{1.0, 0.75, 0.0} %

\makeatletter
\newcommand{\overrightsmallarrow}{\mathpalette{\overarrowsmall@\rightarrowfill@}}
\newcommand{\overarrowsmall@}[3]{%
  \vbox{%
    \ialign{%
      ##\crcr
      #1{\smaller@style{#2}}\crcr
      \noalign{\nointerlineskip}%
      $\m@th\hfil#2#3\hfil$\crcr
    }%
  }%
}
\def\smaller@style#1{%
  \ifx#1\displaystyle\scriptstyle\else
    \ifx#1\textstyle\scriptstyle\else
      \scriptscriptstyle
    \fi
  \fi
}
\makeatother

\mathchardef\ordinarycolon\mathcode`\:
\mathcode`\:=\string"8000
\begingroup \catcode`\:=\active
\gdef:{\mathrel{\mathop\ordinarycolon}}
\endgroup

\usepackage{mathtools}

\makeatletter
\newcommand{\@givenstar}{\;\middle|\;}
\newcommand{\@givennostar}[1][]{\;#1|\;}
\newcommand{\given}{\@ifstar\@givenstar\@givennostar}
\makeatother

\newcommand{\action}{{\hat{a}}}
\newcommand{\Actions}{{\hat{\mathcal{A}}}}
\newcommand{\skill}{{\pi}}
\newcommand{\Skills}{{\mathcal{E}}}

\DeclareFontFamily{U} {MnSymbolA}{}
\DeclareFontShape{U}{MnSymbolA}{m}{n}{
  <-6> MnSymbolA5
  <6-7> MnSymbolA6
  <7-8> MnSymbolA7
  <8-9> MnSymbolA8
  <9-10> MnSymbolA9
  <10-12> MnSymbolA10
  <12-> MnSymbolA12}{}
\DeclareFontShape{U}{MnSymbolA}{b}{n}{
  <-6> MnSymbolA-Bold5
  <6-7> MnSymbolA-Bold6
  <7-8> MnSymbolA-Bold7
  <8-9> MnSymbolA-Bold8
  <9-10> MnSymbolA-Bold9
  <10-12> MnSymbolA-Bold10
  <12-> MnSymbolA-Bold12}{}
\DeclareSymbolFont{MnSyA} {U} {MnSymbolA}{m}{n}
\DeclareMathSymbol{\rhookrightarrow}{\mathrel}{MnSyA}{48}

\usepackage[
  backend=biber
  ,giveninits=true                  %
  ,url=false, isbn=false, doi=false %
  ,style=ieee
]{biblatex}
\AtEveryBibitem{ %
  \clearfield{eprint}
  {}
}
\AtEveryBibitem{                      %
  \iffieldequalstr{eprinttype}{jstor}
  {\clearfield{eprint}}
  {}
}
\renewrobustcmd*{\bibinitdelim}{\,} %
\author{
  Tin Lai\\ %
  School of Computer Science\\
  University of Sydney\\
  Sydney, Australia \\
  \texttt{oscar@tinyiu.com} \\
}

\title{%
  Discover Life Skills for Planning with Bandits via Observing and Learning \\ How the World Works
}

\newcommand{\shortheadtitle}{Discover Life Skills via Learning How the World Works}

\begin{document}

\maketitle

\begin{abstract}
  We propose a novel approach for planning agents to compose abstract skills via observing and learning from historical interactions with the world.
  Our framework operates in a Markov state-space model via a set of actions under unknown pre-conditions.
  We formulate skills as high-level abstract policies that propose action plans based on the current state.
  Each policy learns new plans by observing the states' transitions while the agent interacts with the world.
  Such an approach automatically learns new plans to achieve specific intended effects, but the success of such plans is often dependent on the states in which they are applicable.
  Therefore, we formulate the evaluation of such plans as infinitely many multi-armed bandit problems, where we balance the allocation of resources on evaluating the success probability of existing arms and exploring new options.
  The result is a planner capable of automatically learning robust high-level skills under a noisy environment; such skills implicitly learn the action pre-condition without explicit knowledge.
  We show that this planning approach is experimentally very competitive in high-dimensional state space domains.
\end{abstract}

\section{Introduction}

An agent must learn to reason with its action when it has no prior knowledge of how the environment operates, for example, when it has access to a set of actions without knowing its conditions.
Traditional approaches use deductive reasoning to deduce the likely conditions given the observation~\autocite{russell2002artificial}.
However, such an approach is not effective in an inherently uncertain environment.
In typical robotic applications, sensory noise or environmental factors are often out of the agent's control but contribute to the changes that the agent observes.
For example, a robot agent typically knows the intended effect of its own action (e.g. getting a can of cola), but the action has possibly \emph{unknown} pre-conditions (e.g. conditioned on a coin had been inserted into the vending machine beforehand).
If the action pre-conditions are not given to the agent, the agent needs to reason on its actions by observing how the world works and learning to generate a sequence of actions that conform to its internal beliefs.
Its internal beliefs also need to be robust against noise as \emph{``things might not always work''} due to malfunctions or some confounding factors.

Reinforcement learning in robotics typically concerns the lower-level control of primitive joint movements or motor controls.
Mapping low-level control to the agents' episodic objectives can often learn a policy that performs well among environments.
However, most existing techniques do not scale well in high-dimensional spaces and often ignore the hierarchical structure of the exploitable tasks.
In contrast, we are interested in learning high-level composable skills that are robust against the current state of the environment.
Each skill represents a composable block of knowledge that other skills can reuse.
Telling a robot to \emph{make a cup of tea} might involve the sequence of getting a cup, grabbing a teabag from the cupboard, preparing hot water, etc.
However, when there exists no teabag in the cupboard, then the \emph{make a cup of tea} skill should return a plan that involves heading to shops to replenish teabag stocks.
Alternatively, the boiling steps could be skipped if the kettle is already filled with hot water.
As such, each high-level skill encapsulates the sequence of inner skills to accomplish some tasks and implicitly defines a hierarchical structure.
The analogy of learning life skills through observation is similar to how infants learn skills.
Without prior knowledge, we discover the intent of different primitive actions and the pre-conditions behind those actions.
We slowly accumulate life skills through trial and error and refine our skills as we gather more experience with our skills.
Encapsulating actions as skills help us decompose challenging scenarios and apply what we have learned from the past.
Therefore, we can better account for uncertainty and even be robust against noisy environments as our life-long skills can instinctively provide us with a contemplative plan that best suits the current situation.

This paper proposes a radically different approach than the typical setups in \emph{learning to plan}.
We introduce an encapsulation process under the Markov framework.
Rather than explicitly reasoning on actions' conditions (which is difficult in high dimensional space with noisy observation),
our setup learns high-level policies by capturing the historical successful action trajectories as arms under the Multi-armed Bandit framework.
Our proposed framework---\emph{Markov Encapsulation Process (MEP)}---learns composable skills through successful historical trajectories without explicit reasoning on action conditions.
Each learned skill is specialised in a pre-determined effect and adapts to the current environment by conditioning on the current state.
Each skill is reusable by other skills. Therefore, this approach decomposes planning in a complex environment into one where the agent learns composable skills.
An agent learns life skills through combining historic action sequences that were successful.
Each skill represents some desired effect that the agent wants to achieve, and the realisation of skills is conditioned on the current state.
We formalise the problem setup, provide a simple upper confidence bound algorithm to learn life-long skills, and experimentally demonstrate its effectiveness.

\section{Related works}\label{sec:related_work}
Infinitely many multi-armed bandit~\autocite{wang2008infinitely} is a setup where the number of arms is larger than the possible number of experiments for exploration and exploitation~\autocite{lai2018_BalaGlob,lai2020_BayeLoca,bayati2020unreasonable}.
This setup generalises the approaches in a lifelong planning setting.
Planning by reasoning on the effects and conditions of actions includes classic planners like STRIPS~\autocite{fikes1971strips}.
Typical planners produce sequential plans using known action effects and conditions. Hierarchical Task Networks (HTN) improve on STRIPS by providing a hierarchical alternative to generate plans~\autocite{sacerdoti1975nonlinear,erol1996hierarchical,erol1994umcp,nau1999shop}.
However, these planners necessitate skill hierarchy expert knowledge.
More work has aimed to extend HTN planners to learn the skill hierarchy automatically~\autocite{nejati2006learning,lai2020_RobuHier}. However, skills hierarchy learning requires experts to demonstrate solutions to the problem.

The field of reinforcement learning (RL) has also produced much work on sequential decision making~\autocite{sutton2018reinforcement}, mainly focusing on cases with unknown transition dynamics.
When transition dynamics are learned from observed transitions, model-based RL methods~\autocite{sutton1991dyna} can generate optimal plans maximising rewards. These methods often leverage classic planners such as Monte-Carlo Tree Search~\autocite{coulom2006efficient}, Rapidly-exploring Random Trees~\autocite{lai2021_AdapExpl} for task-based manipulations~\autocite{lai2022_LTR}, or model predictive control~\autocite{williams2017information} to simulate possible futures using the learned transition model.
Related tree-based approaches can also utilise learning-based methods to bootstrap planning~\autocite{lai2021_PlanLear} or to learn kinodynamic constrains~\autocite{lai2022_L4KDLear}.
However, minor errors in learned dynamics---or even noise---compound when planning over long horizons, making such approaches less robust.
RL has also benefited from the concept of hierarchical planning~\autocite{sutton1999between,dietterich2000hierarchical}.
While the nature and number of options are highly dependent on state-space geometry~\autocite{csimcsek2005identifying}, desired state-space manipulations (i.e. skill effects) are always known in advance, as defined by the action space; this is a stark advantage of skills over options.
A method to learn options by identifying transition data clusters was proposed in~\autocite{bakker2004hierarchical}, though it is limited to a few levels of hierarchy.
Because RL is based on the MDP framework, which defines objectives with a reward function, RL-based methods typically cannot handle multiple or changing goals.
Reasoning about actions and their effects is also the focus of relational RL~\autocite{van2005survey}, where state and action spaces are upgraded with objects and their relations. Following similar ideas, the Object-Oriented MDP framework~\autocite{diuk2008object} models state objects and their interactions.
In robotics, the acting agent can also learn the surrounding environment model~\autocite{lai2022_ParaDiff} from sensors or simulator~\autocite{lai2021_SbpePyth}.
These ideas enable scaling planning to problems with very high dimensions, as long as they can be expressed the planning problem into optimisation objectives.

Our problem setup of learning optimal actions under complex system dynamics is related to optimistic planning \autocite{hren2008optimistic}.
An extension that operates in a stochastic environment is investigated in open-loop optimistic planning (OLOP) \autocite{bubeck2010open}, which attempts to search for near-optimal actions within budget strategically.
OLOP operates optimistically with unknown dynamics and assigns a budget to attain minimax optimality.
Our proposed algorithm operates at a higher-level skill where each bandit's arm encapsulates an optimistic policy with the agent's current knowledge about the environment.
It is possible to decompose a complex decision-making problem into a sequence of elementary decisions in a bandit framework \autocite{munos2014bandits}.
Our Markov Encapsulation Process uses the same idea to formulate our problem setup in a hierarchical bandit approach \autocite{scott2010modern} to learn composable skills effectively.

\section{Markov Encapsulation Process}

Learning high-level skills in a high-dimensional environment helps encapsulate multiple actions into composable policies that other skills can reuse.
This is especially useful when an agent has no access to the underlying condition and must only reason via observing how the world transitions.

\subsection{Formulation}

We shall describe the Markov Encapsulation Process---a novel framework for learning high-level skills by observing and forming plans under the multi-armed bandit formulation.

\begin{definition}[Markov Encapsulation Process]
  A Markov Encapsulation Process (MEP) is a tuple $\langle \mathcal{S}, \Actions, E, T, \mathcal{C} \rangle$ composed of a set of states $\mathcal{S}$, a set of primitive actions $\Actions$ with known effects $e \in E$, a set of action condition $\mathcal{C}$, and an exogenous transition noise function $T$.
  An MEP follows the Markov property, which states that the resulting state of a transition $s'$ only depends on the starting state $s$, primitive action $\action \in \Actions$, and the unknown noise function $T$.
\end{definition}

Primitive actions $\Actions$ are how an agent interacts with the world.
The success of any given action is dependent on the action condition $c_\action\in\mathcal{C}$, which is unknown to the agent.

\begin{definition}[Primitive Action]\label{def:action}
  Each primitive action $\action \in \Actions$ is associated with some condition $c_\action \in\mathcal{C}$ in which executing $\action$ would be successful.
  Successfully executing $\action$ in a state in $\mathcal{S}_{\action}$ would apply the action's effect $e$ to the current state.
\end{definition}

\begin{definition}[Condition]\label{def:condition}
  Action condition $c_\action$ is a (potentially empty) set of unit conditions $\{(d_1,v_1),\ldots,(d_m,v_m)\}$ where $d_i$ denotes the dimensional index and $v_i:\mathbb{R} \to \{\mathtt{True},\mathtt{False}\}$ denotes a boolean expression that evaluates on the dimensional value.
  We say that an action is successful if each unit condition in $c_\action$ is met in $s$ and fails otherwise.
  Formally,
  $c_\action$ is fulfilled in $s$ if and only if
  \begin{equation}
    \bigwedge_{i=1}^m v_i(s^{d_i} ) = \mathtt{True},
  \end{equation}
  where $s^{d_i}$ denotes indexing state $s$ at the $d_i$\textsuperscript{th} dimension and extracting its dimensional value.
\end{definition}

\begin{definition}[Effect]\label{def:effect}
  Action $\action$ with effect $e$ modifies the property of state $s$ into a different state in $\mathcal{S}$.
  Effect $e$ is a set of unit effects represented by a tuple of dimensional index and lambda expression, $\{(d_1,M_1),\ldots,(d_m,M_m)\}$, where $M_i$ denotes a lambda term that operates on the dimensional value.
  Applying effect $e$ transitions $s$ to $s'$ where
  \begin{equation}
    \forall i \in \{1,\ldots,m\},\; s'^{d_i} = (\lambda x.M_i~s^{d_i}).
  \end{equation}
\end{definition}
\vspace{.5em}

\begin{definition}[Stochastic Transition Function]
  The transition noise function $T$ is a stochastic process always applied to the state transition before the agent can observe its action's outcome.
  The noise is exogenous which is out of control and out of the agent's awareness.
  An agent can apply $\action_e$ to state $s$ and transitions to state $s'$, where
  $\oplus$ denotes applying effect $e$ to state $s$ and
  \begin{equation}
    \label{eq:state-transition}
    s' = \begin{cases}
      T(s \oplus e) & \text{ if $c_{\action_e}$ is fulfilled}, \\
      T(s)          & \text{ otherwise}.
    \end{cases}
  \end{equation}
  $T$ implicitly defines an underlying probability distribution unknown to the agent.
\end{definition}

\newcommand*\arm{\xi}
\newcommand*\Arms{\Xi}

\subsection{Learning Skills As Bandits}

MEP has no access to the underlying action conditions and must learn high-level skills by observing and inferring how the world behaves under non-deterministic state transitions.
MEP learns to plan by observing action effects and reasoning by encapsulating historical action trajectories as MAB arms $\arm$.
Instead of directly reasoning with action conditions under uncertainty, MEP encapsulates multiple primitive actions and other policies as a high-level skill $\skill_e$.
Each skill has an intended effect $e$, which denotes the desired effect after executing that skill.
Therefore, after observing a sequence of actions achieving a particular effect $e_i$, MEP associates that sequence, as an arm $\arm$, to the skill $\skill_{e_i}$.
Each arm is associated with some skills, and we formulate the arm selection process as an infinitely many multi-armed bandit problem~\autocite{carpentier2015simple} because more arms are created as we observe more how the world behaves.
Therefore, an MEP agent must balance exploring new arms and exploiting the current best set of arms under uncertainty.

\begin{definition}[High-level Skill]
  A MEP skill $\skill_{e} \in \Skills$, identified by its intended effect $e$, is a policy that returns an arm $\arm \in \Arms$.
  The set of arms $\Arms$ is initially composed of primitive actions that contain effect $e$, and MEP learns new arms through reasoning with its observations.
  An arm is a tuple of actions and policies that is learned from experience, and we can query the skill's policy to obtain an arm by conditioning on a state, $\arm \sim \skill_{e}(s)$.
\end{definition}

\subsubsection{Creating new arms from historical actions}
MEP creates an arm via observing historical trajectory to reason on action conditions.
After executing action $\action_e$ and observing the state transitions from $s$ to $s'$, MEP can infer the action's \emph{successfulness} by reasoning on whether the effect $e$ is present in $s'$ (given that $e$ was not already present in $s$).
Note that the presence of (or the lack thereof) $e$ is not indisputable proof of the action being \emph{successful} because the exogenous noise can cause the transitions.
However, MEP can refine its arms by formulating its reasoning under the bandit framework.
Let us consider a concrete example:
\providecommand{\tmpaction}{
  \overbrace{
    \action_1,
    \skill_2,
    \skill_3
  }^{\arm_j}
}
\begin{align}
  \textit{Execution: }\{
   & \rlap{\ensuremath{\overbrace{
        \phantom{\tmpaction , \action_4, \skill_5}
      }^{\arm_k}}}
  \tmpaction,
  \action_4,
  \skill_5,
  \ldots,
  \action_m
  \} \label{ex:executing-actions}  \\
  \textit{Observation: }\{
   & s_1,
  \hspace{.1em} s_2,
  \hspace{.1em} s_3,
  \hspace{.1em} s_4,
  \hspace{.1em} s_5,
  \ldots,
  s_m
  \}                               \\
  \textit{Inference: }\{
   & \texttt{T},
  \hspace{.45em} \texttt{F},
  \hspace{.45em} \texttt{T},
  \hspace{.45em} \texttt{F},
  \hspace{.45em} \texttt{T},
  \hspace{.45em} \ldots,
  \texttt{F}
  \}.
\end{align}
\let\tmpaction\undefined
For now, let us treat the skills $\skill_{i}$ for $i=1,\ldots,m$ as blackbox actions because we have yet to define them fully.
The example begins at some origin state $s_0$, where the agent executed $\action_1$ and observed the state transitions to $s_1$.
The agent then concluded that $\action_1$ is successful (denoted as \texttt{T}) because $\action_1$'s effect exists in $s_1$.
MEP then continued to execute the blackbox skill $\skill_2$, but its intended effect was not observed which leads to the conclusion of $\skill_2$ \emph{failed} at state $s_2$ (denoted as \texttt{F}).
Similar procedures repeat until the end of the horizon.

During the episode, MEP can encapsulate its historical action trajectory as a new arm $\arm$ for their respective effect $e$.
In the example given in~\cref{ex:executing-actions}, $\arm_j$ and $\arm_k$ are two created arms that contain a different set of actions/policies from the same horizon.
Moreover, MEP can actively learn from its observations.
Let $\mathbf{e}_{s_0 \mapsto s_3}=\{e_1,\ldots,e_n\}$ be the set of all unit effects observed from the state transitions of $s_0$ to $s_3$.
The arm $\arm_j$ will be added to each skill $\skill_{e_i}$ for all observed effect $e_i\in\mathbf{e}_{s_0 \mapsto s_3}$.
The same goes for $\arm_j$ of adding to $\skill_i$ for all $e_i\in\mathbf{e}_{s_0 \mapsto s_5}$.

\subsubsection{Executing high-level skill}
Previously we had considered skills as blackbox actions.
Executing skill $\skill_e$, in fact, means querying an arm by conditioning on the current state and executing the arm.
For example, we draw an arm $\arm \sim \skill_2(s_1)$ in~\cref{ex:executing-actions} and then we sequentially execute each nested actions within $\arm$.
The arm might expand to actions $\arm = \{ \action_{\arm.1}, \action_{\arm.2}, \ldots, \skill_{\arm.n}\}$; and the actual observation, starting from $s_1$, is $\{ s_{1.1}, s_{1.2}, \ldots, s_{1.n}\}$ where $s_{1.n} \equiv s_2$ ($s_2$ is the final state after executing $\arm$).
From MEP agent's point of view, it views $\skill_2$ (and the arm $\arm$ that it draws) as a blackbox where the number of \emph{nested} transitions between $s_1$ to $s_2$ (after executing $\action_2$) does not matter.
MEP uses arms to encapsulate multiple actions and only pays attention to whether the desired effect of skill $\skill_2$ is fulfilled in the final state.
After observing the \emph{successfulness} of each arm, we then update our related policies based on the observed results.

\begin{figure*}[tb]
  \centering
  \includegraphics[width=\linewidth]{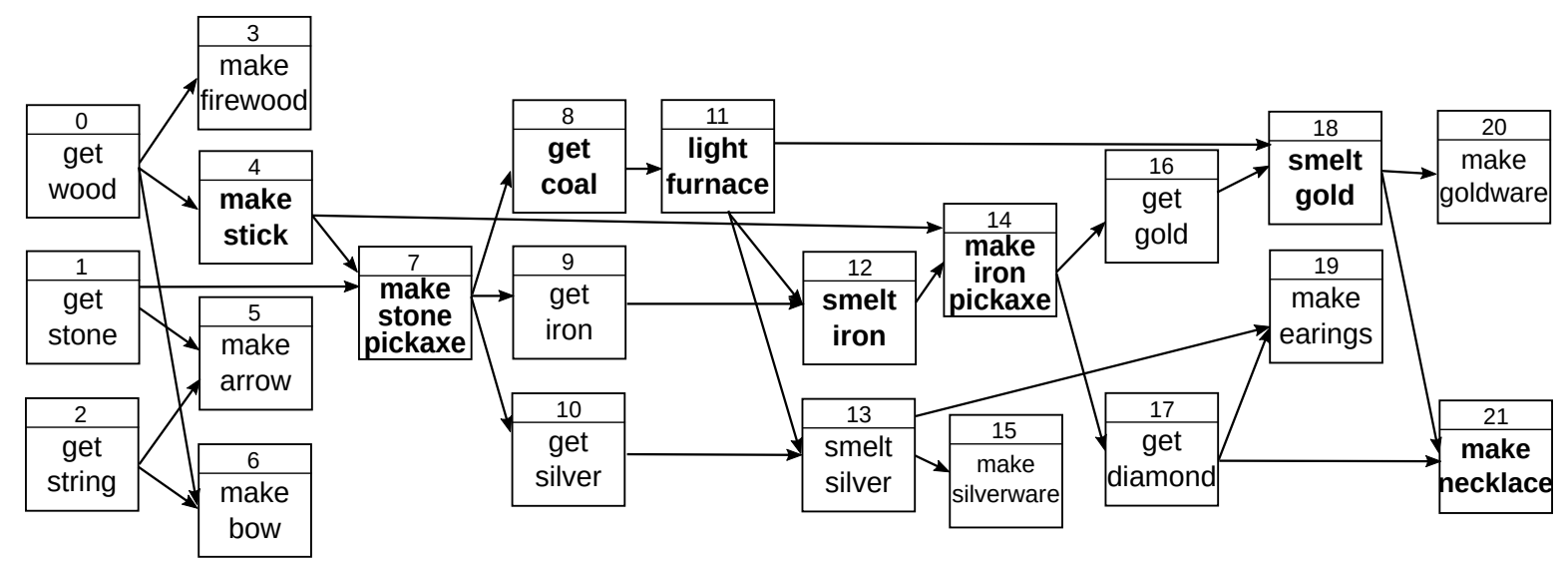}
  \caption{
    Illustration of the dynamics for Mining domain. Nodes are actions and incoming edges are success pre-condition.
    \label{fig:mining-domain}
  }
\end{figure*}

\subsubsection{Modelling arm success probability}

Each arm is created from a historical trajectory under different states; hence, each $\arm\in\skill$ might exhibit different success probability under different circumstances.
Therefore, we model our MAB arm selection process of policy reward as a conditional probability distribution regarding the current state.

One possible difficulty in learning this conditional probability is the high dimensionality of the state space.
In our setup, we utilise Support Vector Machine (SVM) \autocite{cortes1995support} to tackle the dimensionality issue.
SVM creates hyperplanes of the dataspace and attempts to separate each projected state into its respective classes via support vectors.
SVM is effective in high dimensional spaces~\autocite{ghaddar2018high} where the number of features is greater than the number of observations
because it implicitly captures critical dimensions from the high-dimensional data.
The prediction of SVM is given by
\begin{equation}
  f(x) = h(x) + b
\end{equation}
where
\begin{equation}
  h(x) = \sum_i y_i \alpha_j k(s_i, s),
\end{equation}
which lives in a Reproducing Kernel Hilbert Space (RKHS) $\mathcal{F}$ induced by a kernel $k$~\autocite{wahba1999support}.
We use radial basis function for kernel $k$, and train with the loss function
\begin{equation}
  \mathcal{L} = C \sum_i (1 - y_i f(s_i)) + \frac{1}{2} || h ||_\mathcal{F}
\end{equation}
which minimise the error of misclassification rate and the norm of $h$ to produce a sparse machine.
The sparse classifier property is beneficial to our setting as it automatically captures the importance features among our high dimensional settings.
We can then use Platt scaling~\autocite{platt1999probabilistic} to calibrate the probability likelihood of each predicted class for our candidate arm.
We can write the predicted probability of our arm $\arm_j$ being successful under state $s_i$ as
\begin{equation}\label{eq:platt-scaling}
  \mathbb{P}_{\arm_j}(success \mid s_i) = \frac{1}{1 + \exp{(\alpha \cdot f(s_i) + \beta)}},
\end{equation}
where $\alpha,\beta\in\mathbb{R}$ are scalar parameters that are learned by maximum likelihood method across the observations.
The estimation in~\cref{eq:platt-scaling} acts as a logistic transformation of the original score, which can be used to estimate our arm's success probability conditioned on the current state.

\subsubsection{Selecting an arm from skill}

Our MAB arm selection is an Upper Confidence Bound (UCB) based algorithm~\autocite{auer2010ucb} where we actively balance exploration and exploitation to gather more knowledge about the current best arms.
Selecting an arm  $\arm^* \sim \skill(s_i)$ at state $s_i$ is given by
\begin{equation}\label{eq:ucb}
  \arm^* = \argmax_{\arm_j\in\Arms_\skill}
  \left[
    \frac{
      \mathbb{P}_{\arm_j}^{s_i}
      \cdot
      N_s(\arm_j)
    }{
      || \arm_j || \cdot N_t(\arm_j)
    }
    + \gamma \sqrt{\frac{\log \sum_{\arm_k\in\Arms_\skill} N_t(\arm_k) }{ N_t(\arm_j)}}
    \right]
\end{equation}
where $\mathbb{P}_{\arm_j}^{s_i} := \mathbb{P}_{\arm_j}(success \mid s_j)$, $N_s(\arm)$ denotes the number of past successes, $N_t(\arm)$ denotes the number of tries in the past, $||\arm||$ denotes the arm's length, and $\gamma$ is the confidence value that controls the level of exploration.
The UCB approach learns skills' policy by testing arms that haven't been evaluated very often to minimise uncertainty while minimising regret by exploiting the best arm at the current state $s_i$ if it is deemed profitable.

\begin{table}[tb]
  \centering
  \caption{Properties of environments tested in experiments
    \label{tab:env_description}}
  \begin{tabular}{@{}lcccccc@{}}
  \toprule
  \multicolumn{1}{c}{Env.} & \begin{tabular}[c]{@{}c@{}}Consuming\\ effect\end{tabular} & \begin{tabular}[c]{@{}c@{}}Ep.\\ length\end{tabular} & \begin{tabular}[c]{@{}c@{}}Num.\\ actions\end{tabular} & \begin{tabular}[c]{@{}c@{}}Num.\\ nodes\end{tabular} & \begin{tabular}[c]{@{}c@{}}Avg. edges\\ per node\end{tabular} & \begin{tabular}[c]{@{}c@{}}State space\\ size\end{tabular} \\ \midrule
  Mining                   & No                                                         & 40                                                   & 22                                                     & 22                                                   & $2.27\pm1.14$                                                 & $4.19\times10^{6}$                                         \\
  MiningV2                 & Yes                                                        & 80                                                   & 22                                                     & 22                                                   & $3.18\pm1.97$                                                 & $4.19\times10^{6}$                                         \\
  Baking                   & No                                                         & 60                                                   & 30                                                     & 30                                                   & $2.00\pm0.37$                                                 & $1.07\times10^{9}$                                         \\
  Random                   & No                                                         & 100                                                  & 100                                                    & 100                                                  & $1.32\pm0.71$                                                 & $1.27\times10^{30}$                                        \\ \bottomrule
\end{tabular}
\end{table}

\subsubsection{Refining arms}
Since arms are created in a stochastic approach, unnecessary actions often lie within an arm $\arm$.
We can instead randomly mutate an arm to $\arm'$ with some probability $\epsilon$ during the arm selection process.
The mutation randomly prunes away some nested actions within $\arm$ and stores it as a new arm $\arm'$.
This refinement might introduce a broken arm that does not achieve the initially intended effect $e$ (which might benefit some other skill with effect $e'$).
In the event of proper pruning, $\arm'$ would be preferable in~\cref{eq:ucb} than $\arm$, over the long run, due to $\arm'$'s shorter length.

\subsubsection{Tractability}
Our framework attempts to learn high-level policies from historical trajectories.
Therefore, the set of arms $\Arms$ will grow as the agent gathers more experience from episodes.
While more arms can provide our learning approach with more training examples, numerical tractability is also a concern.
Therefore, we employ the following strategies to add historical trajectories as arms.
Arm $\arm_i$ will not be added to $\Arms$ if another semantically identical arm already exists in $\Arms$.
Instead, we will reject such an arm and merely accumulate the existing arm's probability and statistics.
Moreover, we do not wish to continuously increase the number of imperfect arms when we already contain sizeable examples.
Therefore, we can utilise our predictive component to infer the potential success probability of some canidate arm $\arm_j$, and reject $\arm_j$ if its marginalised success probability of $\mathbb{P}_{\arm_j}(success \mid s)$ in all $s$ is less than some threshold $\delta \in \mathbb{R}, 0 \le \delta < 1$.
Having a low marginalised probability implies this arm is not helpful for this skill's desired effect.
In practice, setting $\delta=0.1$ allows us to maintain a tractable set of arms while not compromising the quality of arms.

\section{Numerical Results}

We experimentally evaluate the proposed planner MEP against
\emph{Q-Learning}~\autocite{watkins1992q},
\emph{Monte-Carlo Tree Search} (MCTS)~\autocite{coulom2006efficient} and
\emph{Rapidly-exploring Random Trees} (RRT)~\autocite{lavalle1998rapidly} with multiple search horizon settings.

\begin{figure}[b]
  \centering
  \includegraphics[width=\linewidth]{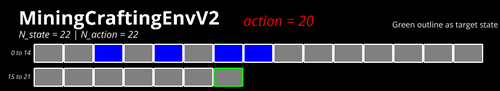}
  \caption{
    Illustration of planning in the \emph{MiningV2} environment.
    Grey and blue represent the dimensional state value being zero and one, respectively.
    Each planner must execute actions to turn the target dimensional state value (outlined in green) from zero to one under noisy transitions.
    \label{fig:miningV2}}
\end{figure}

MCTS is a planning method that builds a non-exhaustive tree of possible futures to search for action with the best long-term score, defined as a reward function of the goal state.
The amount of search is defined by a budget, where we had selected experiments analyse values of $100$, $1000$, and $5000$.
The search is performed every time step after an action is executed. MCTS exploration constant is set to ${1}/{\sqrt{2}}$.
RRT is another sample-based planner which grows an expanding tree from starting state to free space.
RRT can enhance its search by biasing it towards the goal state with low probability ($0.05$).
In our experiments, each tree can expand at most $1000$ and $10000$ nodes.
Q-Learning is an RL-based method that requires training to learn a value function.
We used a discount factor $\gamma=0.99$, a learning rate of $0.1$ and an $\epsilon$-greedy policy with random action probability $0.1$.
A reward of $0$ is given for reaching the goal, and $-1$ rewards are given otherwise. Q-values are tabular and initialised to $0$, ensuring exploration by making the agent optimistic. The method is trained for $20{,}000$ training episodes (typically less than 1 hour).
We report planning success rates, i.e. the number of times the agent manages to reach the given goal. Plan length and planning times are reported for successful runs \emph{only}. All experiments are averaged over $10$ runs on a single core $2.2$GHz.
MEP trains for an episode of length 5,000 to learn a set of sufficient arms before evaluating its performance. In our setup, we use $\epsilon=0.2$ as the probability of pruning and refining the existing arm.
All planners are tested against environments with a noise transition probability of $0.05$, which denotes the probability of one dimension inverting its value every time the agent executes an action, regardless of whether the action is valid or not.

Environmental setup is shown in~\cref{tab:env_description}, which are modelled as directed acyclic dependency graphs, with nodes being state features (i.e. the value $0$ or $1$ of each node is a feature) and edges representing actions and their success conditions.
Environments \emph{Mining}, \emph{Baking} and \emph{Random} all have effects that turn a single state feature from $0$ to $1$.
The environments follow directly from~\autocite{8727471}; however, their proposed planner does not apply to our problem setup as it requires action pre-conditions to be given.
\emph{MiningV2} is a variant of \emph{Mining} with more complex effects, consuming effects, which turn some state features from $1$ back to $0$.
\emph{Mining} and \emph{MiningV2} are modelled after the needs of gathering materials and crafting tools, which in turn are pre-conditions for gathering some other tools.
\emph{Baking} is a simpler environment that directly follows the idea of cooking instructions.
\emph{Random} is a randomly generated environment that contains a complex and chaotic structure.
\Cref{fig:mining-domain} illustrates the overall \emph{Mining} domain dynamics from~\autocite{8727471}.
The MEP's skills learn to capture the implicit hierarchical structure of the overall action dependencies.

Experimental results are shown in~\cref{table:expr-result}.
Overall the proposed method \emph{MEP} consistently outperforms other baselines in most environments, whereas other methods are susceptible to increasing environment complexity and noise.
In particular, methods that do not learn on action conditions fail in complex environments like Random.
Q-learning achieves near-optimal plan length on Mining, as the problem dimension is small enough. However, Q-learning is quickly overwhelmed by increasing problem sizes, and it does not converge in the allocated training time or even runs out of memory.
Similar results are observed in MCTS and RRT; however, the solutions obtained are far from minimal steps.
The two methods also require explicit re-planning to account for state changes and noise at each step.
MEP achieves a short episode horizon in diverse environments due to its approach to encapsulating multiple actions as skills.
The resulting skills can be susceptible to noise, as shown by its degraded performance in complex environments; however, its refinement setup on existing arms can help to enhance further the robustness of its skill under a more comprehensive set of states.

\begin{table*}[tb]
  \centering
  \caption{Experimental results from various methods and environments ($\mu\pm\sigma$ over 10 runs)
    \label{table:expr-result}}
  \resizebox{\linewidth}{!}{%
    \begin{tabular}{@{}ccccccccc@{}}
  \toprule
                                   & \multicolumn{2}{c}{Mining} & \multicolumn{2}{c}{MiningV2} & \multicolumn{2}{c}{Baking} & \multicolumn{2}{c}{Random}                                                                   \\ \cmidrule(l){2-9}
                                   & Success                    & Steps                        & Success                    & Steps                      & Success     & Steps              & Success   & Steps            \\ \midrule
  \multicolumn{1}{c|}{MEP}         & $80.0 \% $                 & $15.65 \pm 1.56$             & $70.0 \% $                 & $ 28.45 \pm 24.21 $        & $100.0 \% $ & $25.6 \pm 5.016$   & $30.0\%$  & $55.9 \pm 9.457$ \\
  \multicolumn{1}{c|}{Q-Learning}  & $50.0 \%$                  & $18.6 \pm 5.1$               & $40.0 \%$                  & $107.2 \pm 68.88 $         & $20.0 \%$   & $43.0 \pm 46.67$   & $0.0\%$   & --               \\
  \multicolumn{1}{c|}{MCTS (100)}  & $10.0 \%$                  & $63.8 \pm 8.7$               & $30.0 \%$                  & $65.43 \pm 4.509$          & $40.0 \%$   & $50.0 \pm 21.38$   & $10.0 \%$ & $47.0 \pm 0.0$   \\
  \multicolumn{1}{c|}{MCTS (1000)} & $10.0 \%$                  & $38.0 \pm 0.0$               & $70.0 \%$                  & $43.33 \pm 22.34$          & $20.0 \%$   & ${27.5 \pm 15.56}$ & $10.0 \%$ & ${25.0 \pm 0.0}$ \\
  \multicolumn{1}{c|}{MCTS (5000)} & $20.0\%$                   & $37.0\pm1.41$                & $70.0\%$                   & $56.71\pm12.92$            & $30.0\%$    & $22.0\pm13.08$     & $0.0\%$   & --               \\
  \multicolumn{1}{c|}{RRT(1000)}   & $60.0 \%$                  & $31.17 \pm 9.30$             & $20.0 \%$                  & $73.5 \pm 9.192 $          & $100\%$     & $33.7 \pm 13.05$   & $0.0 \%$  & --               \\
  \multicolumn{1}{c|}{RRT(10000)}  & $40.0 \%$                  & $31.75 \pm 1.89$             & $50.0 \%$                  & $25.6 \pm 23.37 $          & $100\%$     & $29.0 \pm 1.886$   & $10.0 \%$ & $9.0 \pm 0.0$    \\ \bottomrule
\end{tabular}

  }
\end{table*}

\section{Conclusion}
We proposed an MEP framework and method for planning through encapsulation by dividing the sequential decision-making problem into multiple non-sequential problems with one specific effect each.
Each skill only targets an intended effect and encapsulates the executions under the MAB framework.
The proposed MEP framework is helpful in addressing the high dimensionality and sparse reward issue in RL methods.
Rather than trying to learn a perfect policy that directly brings the agent from the starting state to its target state, the idea of MEP is to decompose the policy and learn smaller but computationally tractable policies that are good at a dedicated skill.
Each skill is responsible for one desired outcome, and when those outcomes are needed, the higher-level policy will hierarchically other well-trained policies.

\printbibliography

\end{document}